# Multi-Head Linear Attention Generative Adversarial Network for Thin Cloud Removal

Chenxi Duan[1] and Rui Li[2, *]

1) The State Key Laboratory of Information Engineering in Surveying, Mapping and Remote Sensing, Wuhan University, 129 Luoyu Road, Wuhan, Hubei 430079, China.

2) School of Remote Sensing and Information Engineering, Wuhan University, 129 Luoyu Road, Wuhan, Hubei 430079, China.

E-mail addresses: chenxiduan@whu.edu.cn (C. Duan), lironui@whu.edu.cn (R. Li)

*Corresponding author.

*Abstract*—In remote sensing images, the existence of the thin cloud is an inevitable and ubiquitous phenomenon that crucially reduces the quality of imageries and limits the scenarios of application. Therefore, thin cloud removal is an indispensable procedure to enhance the utilization of remote sensing images. Generally, even though contaminated by thin clouds, the pixels still retain more or less surface information. Hence, different from thick cloud removal, thin cloud removal algorithms normally concentrate on inhibiting the cloud influence rather than substituting the cloud-contaminated pixels. Meanwhile, considering the surface features obscured by the cloud are usually similar to adjacent areas, the dependency between each pixel of the input is useful to reconstruct contaminated areas. In this paper, to make full use of the dependencies between pixels of the image, we propose a Multi-Head Linear Attention Generative Adversarial Network (MLA-GAN) for Thin Cloud Removal. The MLA-GAN is based on the encoding-decoding framework consisting of multiple attention-based layers and deconvolutional layers. Compared with six deep learning-based thin cloud removal benchmarks, the experimental results on the RICE1 and RICE2 datasets demonstrate that the proposed framework MLA-GAN has dominant advantages in thin cloud removal.

## 1. INTRODUCTION

Beneficial from the rapid advance of Earth Observation technology, substantial remote sensing images with high spatial and spectral resolutions are now available for a varied range of applications including image classification (Lyons et al., 2018; Maggiori et al., 2016), object detection (Li et al., 2017; Xia et al., 2018), and semantic segmentation (Kemker et al., 2018; Zhang et al., 2019a). The revisiting property of orbital acquisitions brings the consecutive monitoring of land surface, ocean, and atmosphere into possibility. As the primary acquisition method, optical sensors play an important part in capturing and generating remote sensing images.

However, optical remote sensing images are inevitably contaminated by cloud cover, which severely impedes the potential of the utilization. Specifically, approximately 35% of the global land surface is obscured by clouds on average at any time (Lin et al., 2013). Clouds retard the transmission of electromagnetic signals, leading to the deficiency of surface information. Thick opaque clouds might obstruct all the echoed signal from the Earth's surface, while thin translucent clouds attenuate the energy of the electromagnetic signals and pollute the covered areas with haze (Li et al., 2019). Since cloud cover is inevitable and widespread which heavily restrain the availability of optical remote sensing images, using appropriate methods to remove the cloud is imperative.

For thick cloud removal, as the land cover information is completely blocked, the primary target is to reconstruct the cloud-contaminated pixels using available information. According to the source of information utilized to reconstruct the cloud-cover regions (Shen et al., 2015), the existing methods can be grouped into spatial information-based methods (Cheng et al., 2017; Shen et al., 2014; Zeng et al., 2013), spectral information-based methods (Li et al., 2012; Malek et al., 2017; Zhang et al., 2002), and temporal information-based methods (Chen et al., 2019b; Ji et al., 2018; Zhang et al., 2020b). Specifically, the spatial information-based methods hammer at restoring the cloudy areas based on the information of the cloud-free area in the image to be repaired. The typical methods include pixel interpolation (Van der Meer, 2012), tensor completion (Liu et al., 2012), and structure-preserving global optimization (Cheng et al., 2017). However, limited by the finite available information, spatial information-based methods can only generate a visually plausible cloud-free image when the cloud cover is very large. For multispectral and hyperspectral images, even though certain bands are contaminated by the cloud, the remaining specific bands can be intact and spotless from pollution. Hence, the spectral information-based methods (Gladkova et al., 2011; Malek et al., 2017) aim at using information from unaffected and unspotted bands to recover lost information in the missing bands. Nevertheless, these methods are invalid when all the bands are polluted. Temporal information-based methods are effective solutions for this predicament, which utilize at least one cloud-free reference image covering the cloud area in the image to be repaired. Residual correction (Zeng et al., 2018), Markov random field (Cheng et al., 2014), principal component analysis (Zhang et al., 2019c), sparsity decomposition (Chen et al., 2019b), and many other advanced technologies have been introduced.

Different from thick cloud removal, the fundamental objective of thin cloud removal is refraining the cloud component and enhancing the surface feature as the information of land cover is not completely obscured by the cloud. Treating thin clouds as the low-frequency component, several methods are proposed by tackling the issue in the frequency domain (Hu et al., 2015; Li et al., 2013). For example, (Liu and Hunt, 1984) proposed the homo-morphic filtering (HF) method by suppressing the thin cloud areas using a low-pass filter. (Shen et al., 2014) conducted and improved the HF method further by treating the thin cloud as the low-frequency information. Due to the high spatial correlation, (Xu et al., 2019) discovered that the highest signal-to-noise ratio trait of the cloud-contaminated data and developed a principal components transform method for thin cloud removal.

Meanwhile, spectral-based methods are also frequently-used for thin cloud removal (Meng et al., 2009; Zhang et al., 2014). The haze optimized transformation (HOT) is a robust and concise method to analyze the properties of the spectral response and haze/cloud spatial distributions of land cover (Zhang et al., 2002). (Chen et al., 2015) further developed an iterative HOT (IHOT) method to address the spectral confusion issue between bright surfaces and haze/clouds. (Lv et al., 2016) proposed an algorithm for cloud removal based on an empirical and radiative transfer model, which is established on the hypothesis that the top of atmosphere (TOA) reflectance from ground targets are linearly related between any two visible bands under clear sky conditions. Although all of these attempts have made encouraging progress and broadened the boundaries of the cloud removal field, the high dependency on specific satellite sensors and hand-crafted parameters restricts the flexibility and adaptability of these traditional methods.

Recently, the convolutional neural network (CNN), a powerful approach to capture nonlinear and hierarchical features automatically from huge volumes of data, has provided a promising and prospective solution to make full use of plentiful data available in the field of remote sensing. The adequate and quality data enable CNN models to deliver fine-grained and high-caliber results which have been successfully and extensively applied to remote sensing image processing (Zhu et al., 2017b), including hyperspectral classification (Li et al., 2020d), object detection (Deng et al., 2018), super-resolution (Jiang et al., 2019), land cover classification (Zhang et al., 2020a) and segmentation (Griffiths and Boehm, 2019).

Accordingly, a great many end-to-end frameworks conducted by CNN have been proposed to remove the cloud component and recover the lost pixels in remote sensing images, especially for thick cloud removal. For example, (Zhang et al., 2018) proposed a unified spatial-temporal-spectral framework based on a deep convolutional neural network (STS-CNN) to reconstruct the contaminated areas. For detecting and removing cloud areas simultaneously, a novel framework using cascaded CNNs was presented by (Ji et al., 2020). (Zhang et al., 2020b) designed a spatio-temporal patch group deep learning framework to remedy the limitations such as the incomplete temporal information of existing methods. Similarly, there are also certain pieces of literature about thin cloud removal. For instance, (Qin et al., 2018) designed a novel deep CNN-based method for removing the thin cloud and dehazing multispectral remote sensing images. Thereafter, (Li et al., 2019) designed an end-to-end framework named residual symmetrical concatenation network (RSC-Net) for thin cloud removal.

Even though the utilization of CNN has demonstrated tremendous potential for cloud removal in remote sensing images, the dependence on cloud-free reference images for recovering restrains the application scenarios of CNN. Actually, due to the time lag in data acquisition, cloud-free reference images from the same region are always neither available nor contain similar surface features. Hence, it is an attractive issue to implement an unsupervised method that does not require reference images.

Promisingly, the generative adversarial network (GAN) provides a workable solution for unsupervised learning on complex distributions. The model generates fake samples via a two-player minimax game between the generative model (G) and the discriminative model (D) (Goodfellow et al., 2014). Subsequently, the paired image-to-image translation (Isola et al., 2017) and unpaired image-to-image translation (Zhu et al., 2017a) using the GAN framework were developed. In (Enomoto et al., 2017), the conditional-GAN was extended to multispectral images for cloud removal. (Singh and Komodakis, 2018) addressed the cloud removal issued via a cycle-GAN-based network. In (Zheng et al., 2020), a two-stage scheme was designed for the single image cloud removal which comprises cloud segmentation stage and image restoration stage. (Li et al., 2020a) tackled the thin cloud removal by combining a physical model of cloud distortion with GAN.

Besides, image dehazing and image deraining are two similar tasks like thin cloud removal in the computer vision community where substantial efforts have been poured in from diverse perspectives. For example, (Zhu et al., 2015) presented a color attenuation prior (CAP) for single image haze removal. (Chen et al., 2019a) proposed an end-to-end gated context aggregation network (GCANet) to recover the haze-free image. In (Wang et al., 2019), a local-to-global framework named Spatial Attentive Network (SPANet) was proposed to remove rain streaks.

Considering the surface features of adjacent areas in a certain region are normally semblable, the relationships between pixels in the image of the input are useful to reconstruct contaminated areas and maintain visual consistency. Benefit from its strong capability to capture long-range dependencies, dot-product attention mechanisms have been successfully introduced into the GAN framework (Zhang et al., 2019b). Whereas, the utilization of the dot-product attention mechanism often comes with significant memory and computational costs, which increase quadratically with the size of the input over space and time. Hence, the pervasive and flexible combination between attention mechanisms and networks remains an intractable problem. In this paper, to remedy this dilemma, we propose a semi-supervised Multi-Head Linear Attention Generative Adversarial Network (MLA-GAN) for thin cloud removal based on our previous studies on reducing the complexity of the attention mechanism (Li et al., 2020b; Li et al., 2020c). The main contributions of the proposed MLA-GAN method are twofold. On the one hand, we demonstrate the effectiveness of the attention mechanism on the thin cloud removal task. On the other hand, we provide a novel framework to combine the attention mechanism and GAN reasonably.

## 2. METHODS

### 1) Dot-Product Attention

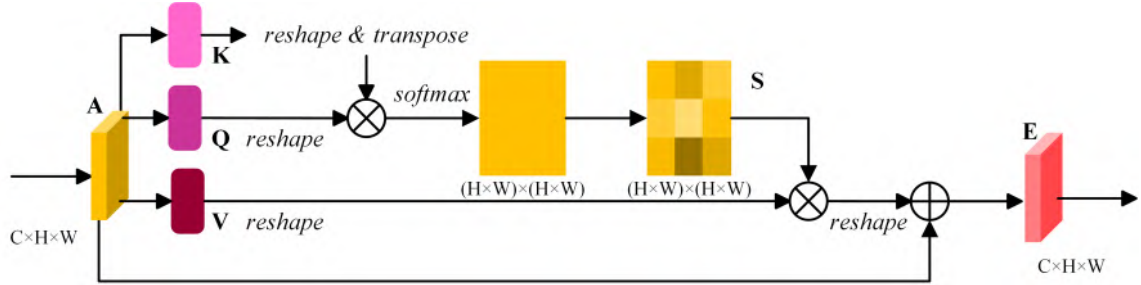

Fig.1 The diagram of the dot-product attention.

Let $N$ and $C$ represent the length of input sequences and the number of input channels respectively, and the input feature is signified as $X = [x_1, \cdots, x_N] \in \mathbb{R}^{N \times C}$. Firstly, three projected matrices $W_q \in \mathbb{R}^{D_x \times D_k}$, $W_k \in \mathbb{R}^{D_x \times D_k}$, and $W_v \in \mathbb{R}^{D_x \times D_v}$ are employed by the scaled dot-product attention to corresponding query matrix $Q$, the key matrix $K$, and the value matrix $V$:

$$\begin{cases} Q = XW_q \in \mathbb{R}^{N \times D_k}; \\ K = XW_k \in \mathbb{R}^{N \times D_k}; \\ V = XW_v \in \mathbb{R}^{N \times D_v}. \end{cases} \tag{1}$$

It is worthy of note that the dimensions of the query matrix and key matrix are supposed to be equivalent and all the vectors in this section are column vectors by default. Accordingly, the similarity between the *i*-th query feature $q_i^T \in \mathbb{R}^{D_k}$ and the *j*-th key feature $k_j \in \mathbb{R}^{D_k}$ is evaluated by a normalization function $\rho$ as $\rho(q_i^T k_j) \in \mathbb{R}^1$. Normally, since the query feature and key feature are calculated using disparate layers, the similarities between $\rho(q_i^T k_j)$ and $\rho(q_j^T k_i)$ are not symmetric. By computing the similarities between all pairs of positions and taking the similarities as weights, the scaled dot-product attention module generates the value at position *i* by aggregating the value features from all positions using weighted summation:

$$D(\boldsymbol{Q}, \boldsymbol{K}, \boldsymbol{V}) = \rho(\boldsymbol{Q}\boldsymbol{K}^T)\boldsymbol{V}. \tag{2}$$

The softmax is a common normalization function:

$$\rho(\boldsymbol{Q}^T\boldsymbol{K}) = softmax_{row}(\boldsymbol{Q}\boldsymbol{K}^T), \tag{3}$$

where $softmax_{row}$ denotes the softmax function is exploited along each row of matrix $\boldsymbol{Q}\boldsymbol{K}^T$.

Modeling the similarities between each pair of targets of the input, the $\rho(\boldsymbol{Q}\boldsymbol{K}^T)$ can thoroughly extract the global dependencies contained in the features. Nevertheless, for $\boldsymbol{Q} \in \mathbb{R}^{N \times D_k}$ and $\boldsymbol{K}^T \in \mathbb{R}^{D_k \times N}$, the product between $\boldsymbol{Q}$ and $\boldsymbol{K}^T$ belongs to $\mathbb{R}^{N \times N}$, leading to the $O(N^2)$ memory and computation complexity. Thus, the high resource-demanding of dot-product critically limits its application on large inputs. Therefore, it is requisite to reduce the huge demand for computational resources of the scaled dot-product attention.

## 2) Generalization and simplification of the dot-product attention mechanism

According to equation (2), the *i*-th row of result matrix generated by the dot-product attention module under the condition of softmax normalization function can be written as:

$$D(\boldsymbol{Q}, \boldsymbol{K}, \boldsymbol{V})_i = \frac{\sum_{j=1}^{N} e^{q_i^T k_j} \boldsymbol{v}_j}{\sum_{j=1}^{N} e^{q_i^T k_j}}. \tag{4}$$

To generalize equation (4) to any normalization function, equation (4) can be rewritten as:

$$\begin{aligned} D(\boldsymbol{Q}, \boldsymbol{K}, \boldsymbol{V})_i &= \frac{\sum_{j=1}^{N} \sin(\boldsymbol{q}_i, \boldsymbol{k}_j) \boldsymbol{v}_j}{\sum_{j=1}^{N} \sin(\boldsymbol{q}_i, \boldsymbol{k}_j)}, \\ \sin(\boldsymbol{q}_i, \boldsymbol{k}_j) &\geq 0, \end{aligned} \tag{5}$$

in which $\sin(\boldsymbol{q}_i, \boldsymbol{k}_j)$ can be expanded as $\phi(\boldsymbol{q}_i)^T \varphi(\boldsymbol{k}_j)$ that estimates the similarity between the $\boldsymbol{q}_i$ and $\boldsymbol{k}_j$. Particularly, equation (5) is equivalent to equation (4) if $\phi(\cdot) = \varphi(\cdot) = e^{(\cdot)}$, whereupon equation (4) can be rewritten as equation (6) and be simplified as equation (7):

$$D(\boldsymbol{Q}, \boldsymbol{K}, \boldsymbol{V})_i = \frac{\sum_{j=1}^{N} \phi(\boldsymbol{q}_i)^T \varphi(\boldsymbol{k}_j) \boldsymbol{v}_j}{\sum_{j=1}^{N} \phi(\boldsymbol{q}_i)^T \varphi(\boldsymbol{k}_j)}, \tag{6}$$

$$D(Q,K,V)_i = \frac{\phi(q_i)^T \sum_{j=1}^{N} \varphi(k_j) v_j^T}{\phi(q_i)^T \sum_{j=1}^{N} \varphi(k_j)}. \tag{7}$$

For $K \in \mathbb{R}^{D_k \times N}$ and $V^T \in \mathbb{R}^{N \times D_v}$, the product between $K$ and $V^T$ belongs to $\mathbb{R}^{D_k \times D_v}$, thereby considerably reducing the complexity of the scaled dot-product attention mechanism. The suitable $\phi(\cdot)$ and $\varphi(\cdot)$ enable the above scheme to achieve the distinguished performance with finite complexity (Katharopoulos et al., 2020; Li et al., 2020c).

### 3) Linear Attention Mechanism

Based on the first-order approximation of Taylor expansion, we proposed an attention mechanism in our previous work (Li et al., 2020b), which is shown as equation (5):

$$e^{q_i^T k_j} \approx 1 + q_i^T k_j. \tag{8}$$

Since the above approximation cannot guarantee to be nonnegative, $q_i$ and $k_j$ are normalized by $l_2$ norm to ensure $q_i^T k_j \geq -1$:

$$sim(q_i, k_j) = 1 + \left(\frac{q_i}{\|q_i\|_2}\right)^T \left(\frac{k_j}{\|k_j\|_2}\right). \tag{9}$$

Therefore, equation (5) can be rewritten as equation (10) and simplified as equation (11):

$$D(Q,K,V)_i = \frac{\sum_{j=1}^{N}\left(1 + \left(\frac{q_i}{\|q_i\|_2}\right)^T \left(\frac{k_j}{\|k_j\|_2}\right)\right) v_j}{\sum_{j=1}^{N}\left(1 + \left(\frac{q_i}{\|q_i\|_2}\right)^T \left(\frac{k_j}{\|k_j\|_2}\right)\right)}, \tag{10}$$

$$D(Q,K,V)_i = \frac{\sum_{j=1}^{N} v_j + \left(\frac{q_i}{\|q_i\|_2}\right)^T \sum_{j=1}^{N}\left(\frac{k_j}{\|k_j\|_2}\right) v_j^T}{N + \left(\frac{q_i}{\|q_i\|_2}\right)^T \sum_{j=1}^{N}\left(\frac{k_j}{\|k_j\|_2}\right)}. \tag{11}$$

The equation (11) can be turned into in a vectorized form:

$$D(Q,K,V) = \frac{\sum_j V_{i,j} + \left(\frac{Q}{\|Q\|_2}\right)\left(\left(\frac{K}{\|K\|_2}\right)^T V\right)}{N + \left(\frac{Q}{\|Q\|_2}\right) \sum_j \left(\frac{K}{\|K\|_2}\right)^T_{i,j}} \tag{12}$$

Since $\sum_{j=1}^{N}\left(\frac{k_j}{\|k_j\|_2}\right)v_j^T$ and $\sum_{j=1}^{N}\left(\frac{k_j}{\|k_j\|_2}\right)$ can be calculated and reused for each query, time and memory complexity of the attention based on equation (12) is $O(N)$.

The validity and efficiency of the proposed attention have been testified through extensive ablation experiments and analysis (Li et al., 2020b).

### 4) Architecture of MLA-GAN

As can be seen from Fig.2, the proposed MAL-GAN consists of two networks: a generative network (G) and a discriminative network (D). When training the network, G and D compete and with each other in a minimax game, while G strives to generate cloud-free and realistic images, and D endeavors to discriminate the plausible images generated by G from the real cloud-free images. This process can be formulated as follows:

$$\min_G \max_D V(G,D) = E_{x\sim p_{data}(x)}[\log(D(x))] + E_{z\sim p(z)}\left[\log\left(1 - D(G(z))\right)\right]. \qquad (13)$$

Here, $x$ denotes the real samples whose distributions are $p_{data}$, $z$ represents the cloud-contaminated images with $p_z$ distribution, and $V(G,D)$ indicates the loss value of G and D. As can be seen from equation (13), G minimizes the $V(G,D)$ by providing high-caliber images, while D maximizes the $V(G,D)$ via distinguishing between real and fake images. After the $V(G,D)$ converges to a Nash equilibrium in the training set, we then utilized the G and cloud-contaminated image for testing. The loss function of MLA-GAN is given as:

$$L = arg \min_G \max_D L_{cGAN}(G,D) + L_{L1}(G), \qquad (14)$$

$$L_{cGAN}(G,D) = E_{x,y\sim p_{data}(x,y)}[\log D(x,y)] + E_{x\sim p_{data}(x),z\sim p_z(z)}\left[\log\left(1 - D(x,G(x,z))\right)\right], \qquad (15)$$

$$L_{L1}(G) = \frac{1}{CHW}\sum_{c=1}^{C}\sum_{h=1}^{H}\sum_{w=1}^{W} \lambda_c \left|I_{gt}^{(w,h,c)} - G(I_{in})^{(w,h,c)}\right|. \qquad (16)$$

where C, H, and W indicate the number of channels, height, and width of the image respectively,

$I_{in}$ is the input cloud image, and $I_{gt}$ is the ground truth.

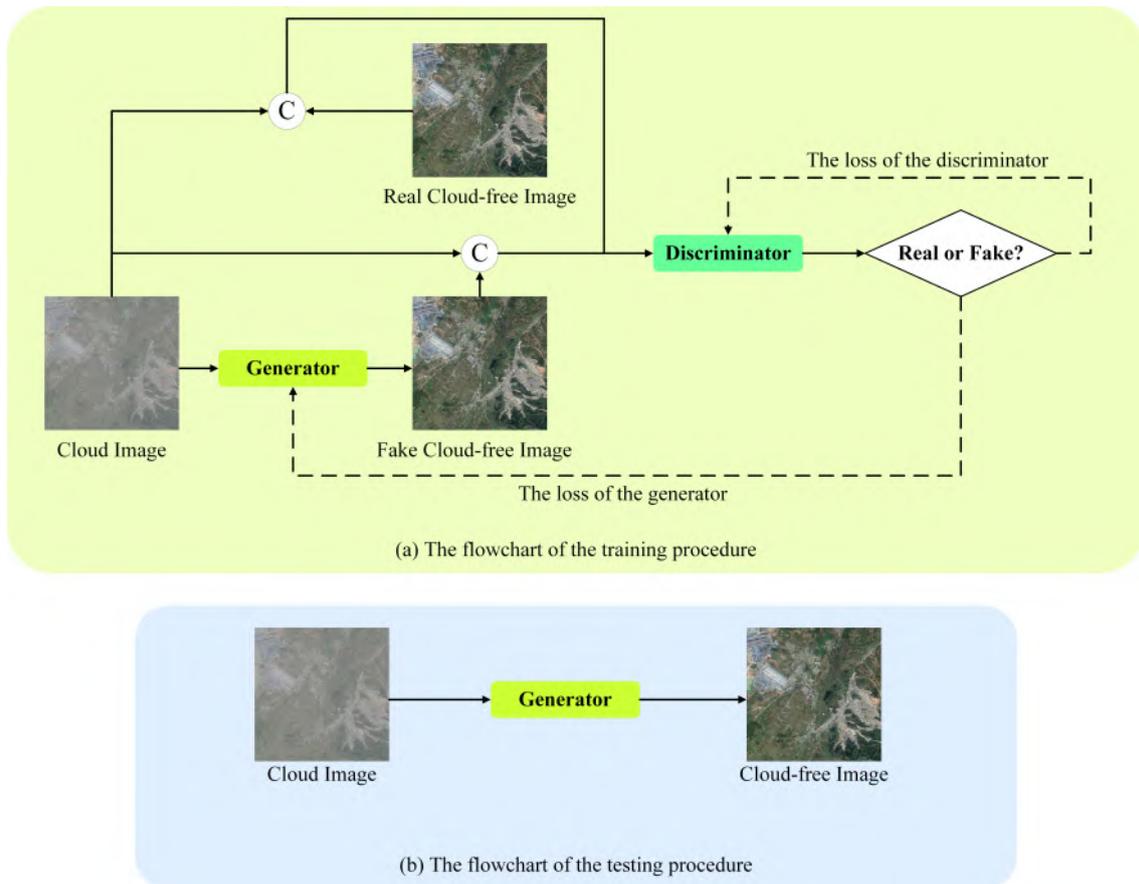

Fig.2 The flowchart of the proposed MLA-GAN.

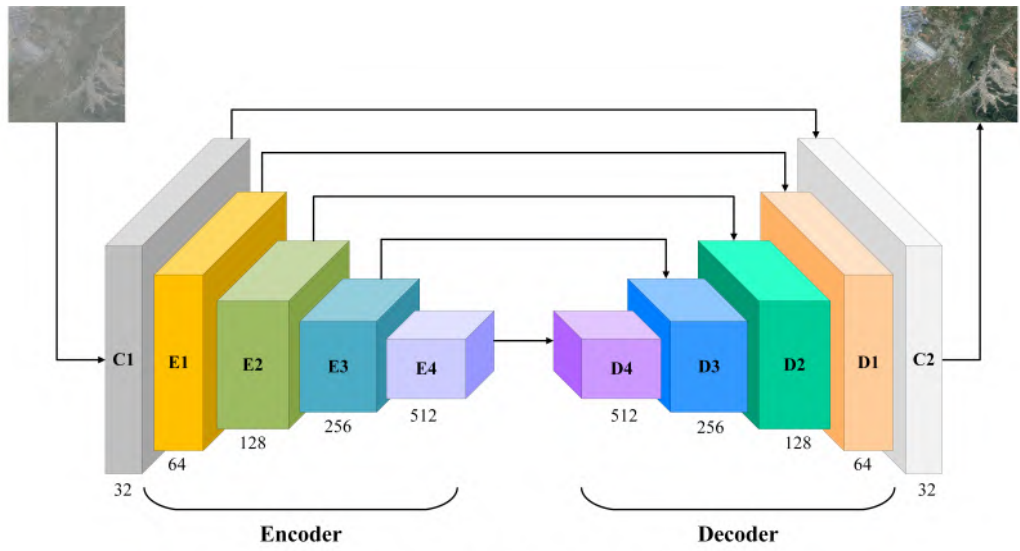

(a) The Integrated Structure of Proposed MAResU-Net

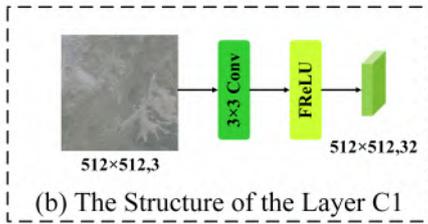

(b) The Structure of the Layer C1

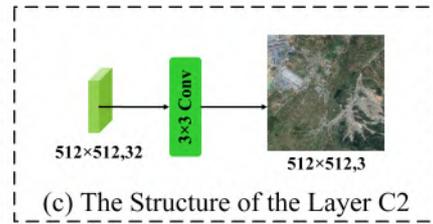

(c) The Structure of the Layer C2

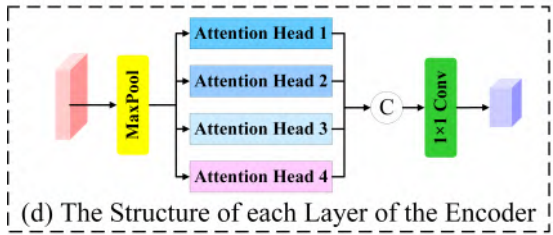

(d) The Structure of each Layer of the Encoder

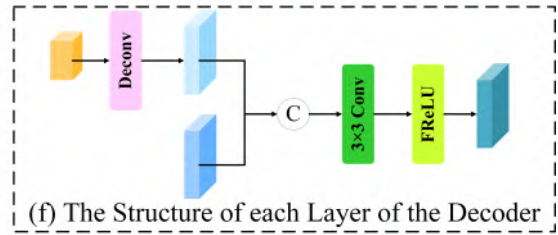

(f) The Structure of each Layer of the Decoder

Fig.3 The structure of the generator.

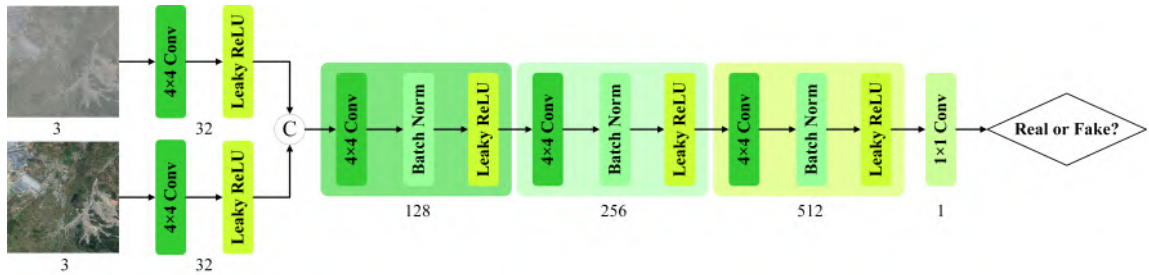

Fig.4 The structure of the discriminator.

## 3. EXPERIMENTAL RESULTS AND DISCUSSION

**1) Datasets**

In this paper, we evaluate the performance of the proposed MLA-GAN with other models on the open-source dataset named Remote sensing Image Cloud Removing (RICE) provided by (Lin et al., 2019). There are two subsets called RICE1 and RICE2 in the RICE1 dataset. To be specific, the RICE1 contains 500 samples collected by Google Earth, where each sample has a cloudy image and the corresponding cloud-free image with 512×512 resolution. The cloudy/cloud-free images are obtained by setting the cloud layer to display or not. The 736 groups of 512×512 samples in the RICE2 are constructed from Landsat 8 OLI/TIRS data by using LandsatLook images with georeferenced in Earth Explorer, where the cloud-free images are manually selected at the same location with a cloud image time less than 15 days. The data samples of RICE are shown in Fig. 5. For each dataset, 80% of samples are selected as the training set, while the remainings are set as the test set.

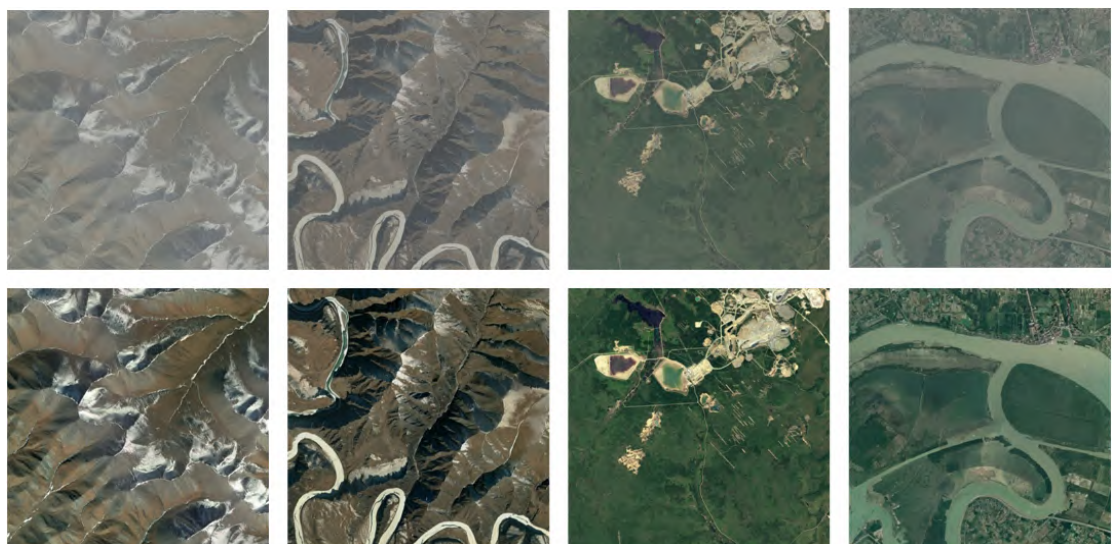

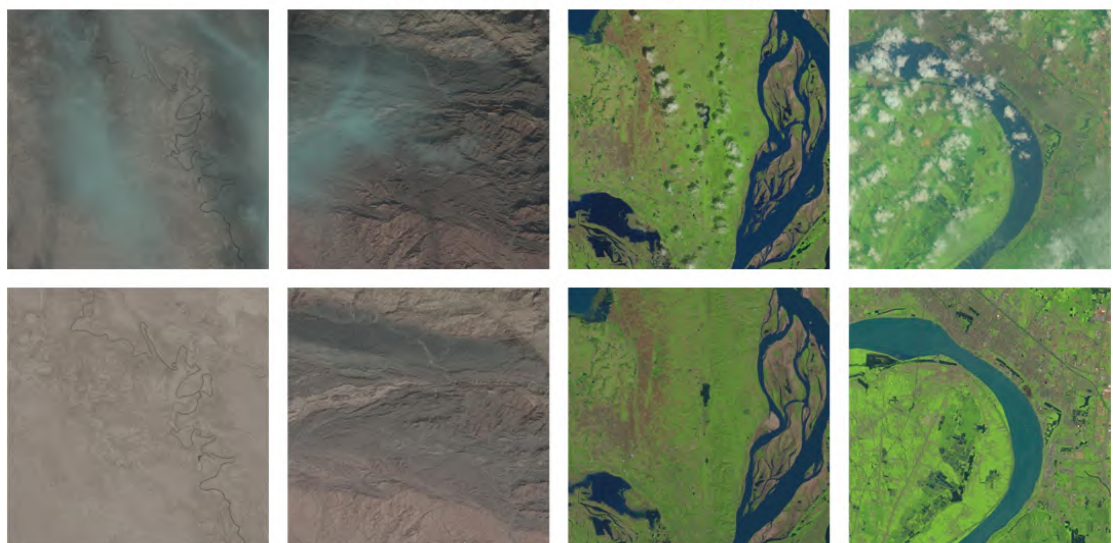

Fig. 5. The data samples of RICE.

## 2) Evaluation Metrics

To measure the quality of the generated cloud-free image, peak signal to noise ratio (PSNR) (Huynh-Thu and Ghanbari, 2008) and structural similarity index (SSIM) (Wang et al., 2004) are utilized in this paper as image quality evaluation metrics to indicate the cloud removal ability.

Specifically, the PSNR denotes the ratio of the maximum pixel intensity to the power of the distortion which can be formulated as:

$$PSNR = 10 \cdot log_{10}\left(\frac{MAX}{MSE}\right) = 20 \cdot log_{10}\left(\frac{MAX}{\sqrt{MSE}}\right) \tag{17}$$

where $MAX = (2^n - 1)^2$ and $n$ is the bits of the pixel value and MSE measures the mean square error between the image $X$ and $Y$:

$$MSE = \frac{1}{HW}\sum_{i=1}^{H}\sum_{j=1}^{W}\left(X(i,j) - Y(i,j)\right)^2 \tag{18}$$

Generally, the value of PSNR situates between 20 to 40, while the larger value denotes the better prediction quality.

SSIM evaluates the similarity between two images through brightness, contrast, and structure:

$$l(X,Y) = \frac{2\mu_X\mu_Y + C_1}{\mu_X^2 + \mu_Y^2 + C_1} \tag{19}$$

$$c(X,Y) = \frac{2\sigma_X\sigma_Y + C_2}{\sigma_X^2 + \sigma_Y^2 + C_2} \tag{20}$$

$$s(X,Y) = \frac{\sigma_{XY} + C_3}{\sigma_X\sigma_Y + C_3} \tag{21}$$

$$SSIM = l(X,Y) \cdot c(X,Y) \cdot s(X,Y) \tag{22}$$

where $C_1$, $C_2$, and $C_3$ are constants for the sake of avoiding divide zero error, $\mu$ and $\sigma$ represent the mean and variance of the image, $\sigma_{XY}$ denotes the covariance of the image $X$ and $Y$.

3) **Experimental Setting**

In order to comprehensively measure the performance of the proposed MLA-GAN, we consider several benchmark comparators including the traditional cloud removal method AHF (Shen et al., 2014), the traditional image dehazing method CAP (Zhu et al., 2015), the CNN-based image dehazing method GCANet (Chen et al., 2019a), the CNN-based cloud removal method RSC-Net (Li et al., 2019), the GAN-based cloud removal method Cloud-GAN (Singh and Komodakis, 2018), and the GAN-based image deraining method SPANet (Wang et al., 2019).

**4) Results and analysis**

TABLE I

QUANTITATIVE ANALYSIS ON RICE1 DATASET

| Method | PSNR | SSIM |
|---|---|---|
| AHF (Shen et al., 2014) | 19.6072 | 0.8899 |
| CAP (Zhu et al., 2015) | 21.1563 | 0.9294 |
| GCANet (Chen et al., 2019a) | 28.8788 | *0.9616* |
| RES-Net (Li et al., 2019) | 28.1012 | 0.9561 |
| Cloud-GAN (Singh and Komodakis, 2018) | 26.1461 | 0.9248 |
| SPANet (Wang et al., 2019) | *29.9704* | 0.9562 |
| MLA-GAN | **32.0487** | **0.9780** |

The generated results of different methods on the RICE1 dataset are shown in Fig. 6 and the quantitative comparisons are demonstrated in Table 1. As virtually all the clouds in RICE1 are thin and the features of the land surface are not completely lost, every method maintains the correct geometric structure and spatial information of the image. Even though nearly all the generated results are visually plausible, the quantitative indicators explicitly illustrate the tremendous gaps between different methods. Lacking the training procedure and depending on the handcrafted features, the performances of AHF and CAP are the worst among the seven methods. The principle of AHF to remove the cloud is suppressing the low-frequency component but enhancing the high-frequency component, however, the low-frequency in cloud-free areas is also eliminated. CAP removes the haze based on the constant scattering coefficient assumption, which tends to underestimate the transmission, especially for inhomogeneous atmosphere conditions. Therefore, the performance of CAP heavily relies on the image acquisition procedure. As can be seen from Fig. 6, CAP fails to remove the thin cloud of the first image.

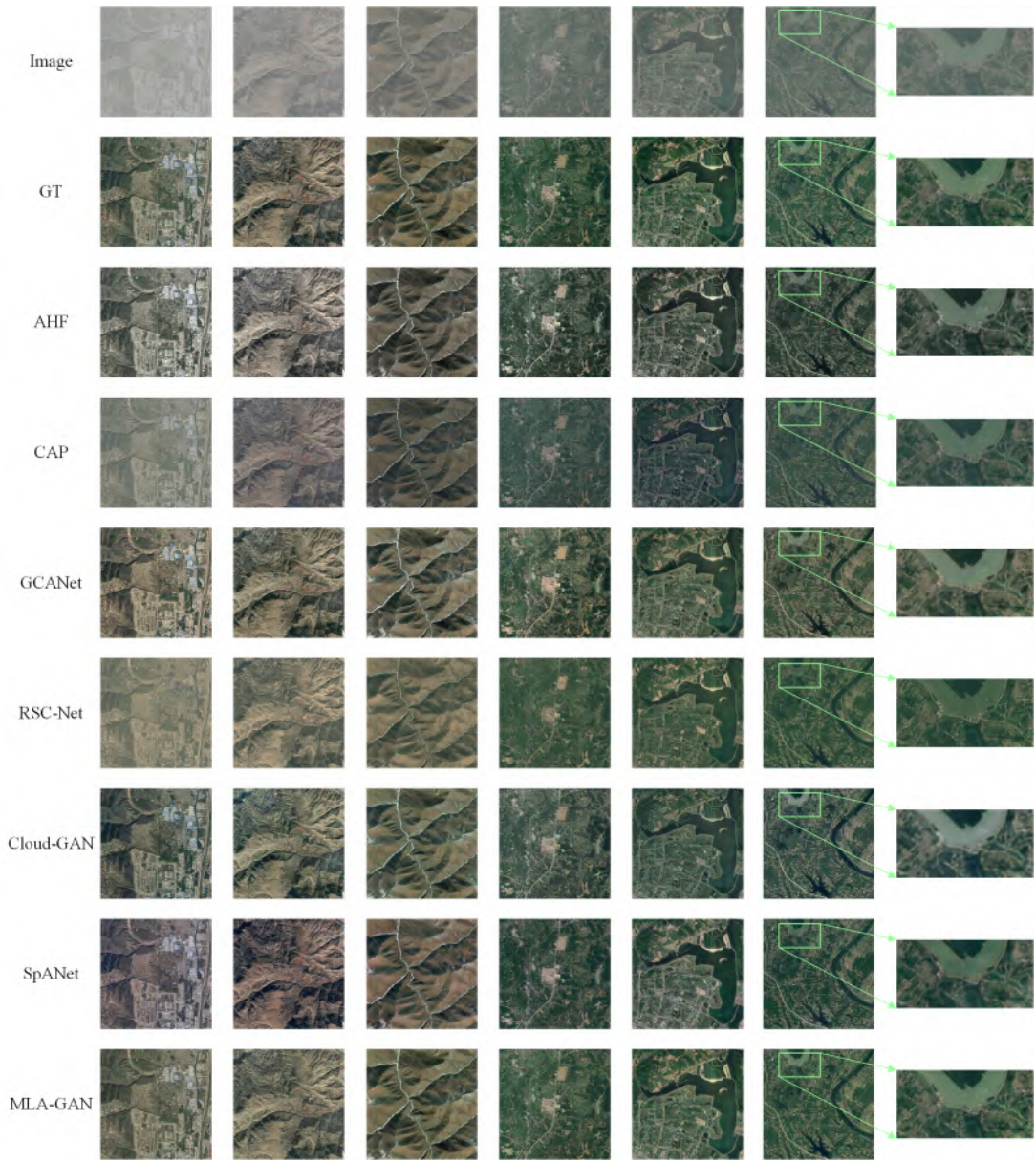

Fig. 6. The generated results on RICE1.

The results show that deep-learning-based methods transcend the traditional methods both visually and quantitatively. The repaired images of Cloud-GAN maintain the major texture and structure features of the background and look visually plausible, however, the enlarged image in the last column illustrates the details are blurry. The reason is that the cycle structure in Cloud-GAN fuses the information contained in clouds to help the network restore cloud-free images. Both GCANet and RSC-Net are CNN-based network architectures that guide models to learn the background features and update the parameters by calculating the discrepancy between inputs cloudy images and the corresponding cloud-free images. Hence, their results surpass those of Cloud-GAN which is relatively fine to retain the textural and structural features of the background, but they differ from reference images in tone. For example, the river color generated by the RSC-Net in the last column is totally altered compared with the ground truth. In SPANet, based on a two-round four-directional recurrent neural network with ReLU and identity matrix initialization (IRNN) a spatial attentive module is harnessed to capture contextual information and generate the cloud masks, which guarantee the performance of SPANet. However, the obvious striations degenerate the quality of the results generated by SPANet. By contrast, the proposed MLA-GAN delivers not only visually reasonable but also quantitatively high-caliber cloud removal results.

Specifically, there are more than 2 dB improvements of the PSNR values and 1% improvements of the SSIM values compared with other methods. Since the long-range relationships between the entire positions in the feature maps are taken into consideration via each layer of the encoder, the texture and structure feature generated by the proposed MLA-GAN retains more coherence and vraisemblance. As can be seen from the zoomed image in the last column of Fig. 6, more fine-grained details are maintained and recovered by MLA-GAN, which keeps the identical color tonality with the ground truth.

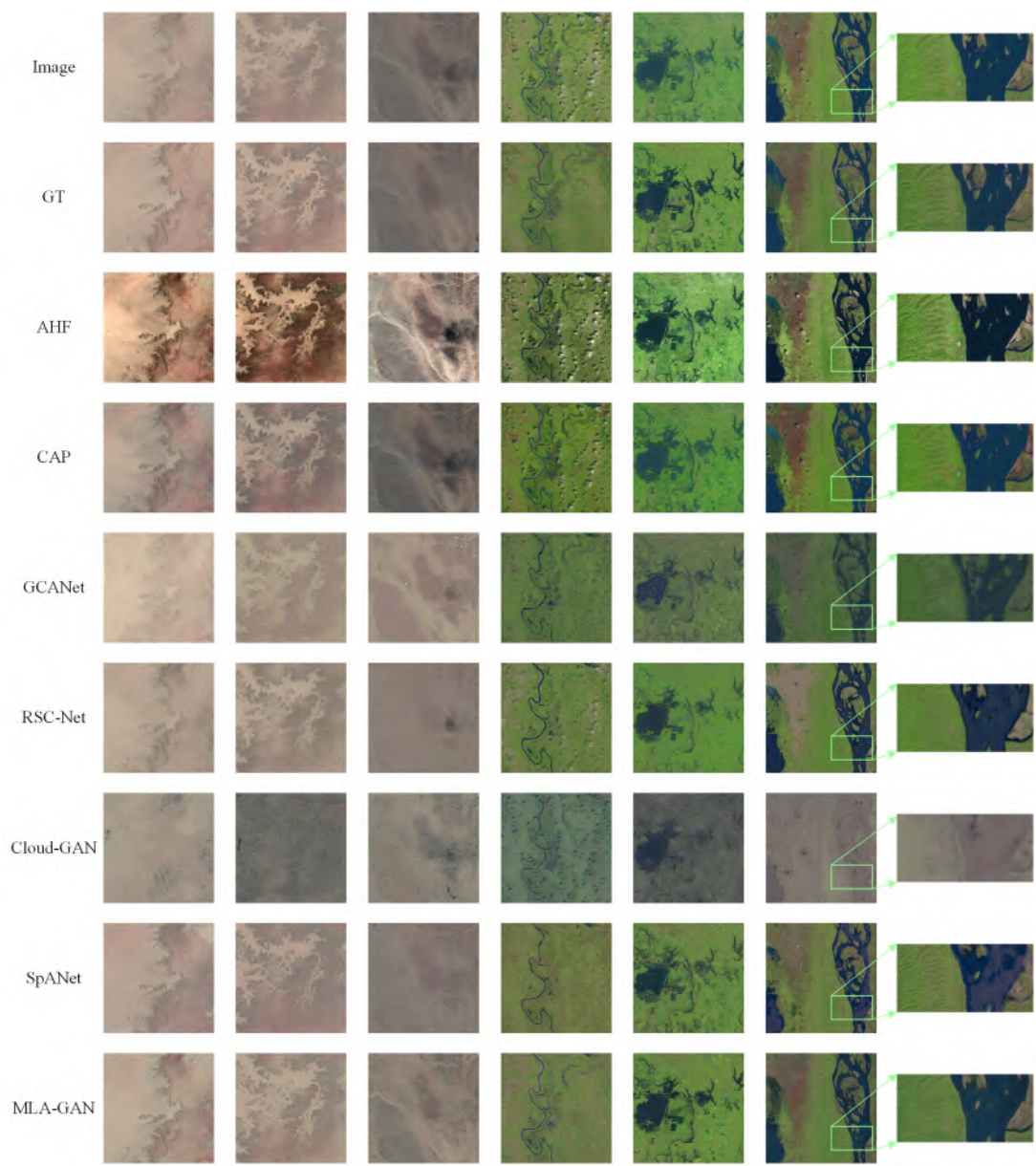

Fig. 7. The generated results on RICE2.

The superiority of the proposed MLA-GAN is more significant when it comes to the results of RICE2. Concretely, as s part of images in the RICE2 dataset are interfered by thick clouds where the surface features in the cloud-contaminated area are completely indistinguishable, the AHF and CAP are incapable of recovering the thick-cloud-coved information since they do not consider the adjacent available characteristics. In contrast, almost all deep-learning-based methods except Clou-GAN achieve visually satisfactory performances. However, the enlarged images in the last column of Fig. 7 dramatically illustrate the enormous disparities between different methods. The result of GCANet is much darker than the original image and round truth. The RSC-Net obliterates some little rocks located in the river, which can be seen from the upper-right in the enlarged image. The color tunes of Cloud-GAN are widely divergent from the raw images. As for SPANet, the generated river area is polluted by the grey panes partly. In comparison, the proposed MLA-GAN retains enough and vivid details while maintains an accordant and accurate hue with the original image. Besides, the quantitive evaluations explicitly demonstrate the preponderance of MAL-GAN. As can be seen from Table II, nearly 4dB improvements in the PSNR and 2.5%

TABLE II

QUANTITATIVE ANALYSIS ON RICE2 DATASET

| Method | PSNR | SSIM |
|---|---|---|
| AHF (Shen et al., 2014) | 18.1159 | 0.7796 |
| CAP (Zhu et al., 2015) | 20.5898 | 0.8523 |
| GCANet (Chen et al., 2019a) | 26.8730 | *0.9173* |
| RESNet (Li et al., 2019) | 26.2788 | 0.9023 |
| Cloud-GAN (Singh and Komodakis, 2018) | 23.5385 | 0.8958 |
| SPANet (Wang et al., 2019) | *27.5042* | 0.8970 |
| MLA-GAN | **31.4991** | **0.9441** |

improvements in the SSIM are brought in.

5) **Ablation study**

In order to verify and analyze the effectiveness of the linear attention mechanism and the number of attention heads on the performance of MLA-GAN, we implement five extra frameworks, i.e., MC-GAN (replacing each layer in the encoder of MLA-GAN with CNN), MLA-GAN1 (setting the head number as one), MLA-GAN2 (setting the head number as two), MLA-GAN6 (setting the head number as six), and MLA-GAN8 (setting the head number as eight). The results indicate that attention-based methods exceed the CNN-based method, due to the mighty capability of the attention mechanism to extract the long-range dependency. Meanwhile, the increments in the number of attention head cannot always yield performance enhancements.

TABLE III

THE ABLATION STUDY

| Method | Head | Num | RICE1 | | RICE2 | |
|---|---|---|---|---|---|---|
| | | | PSNR | SSIM | PSNR | SSIM |
| MC-GAN | ✗ | - | 28.0479 | 0.9697 | 25.4098 | 0.9085 |
| MLA-GAN1 | ✓ | 1 | 28.2773 | 0.9712 | 26.6308 | 0.9187 |
| MLA-GAN2 | ✓ | 2 | 29.9264 | 0.9745 | 29.2265 | *0.9329* |
| MLA-GAN | ✓ | 4 | **32.0487** | *0.9780* | **31.4991** | **0.9441** |
| MLA-GAN6 | ✓ | 6 | *32.1306* | 0.9763 | 29.3004 | 0.9281 |
| MLA-GAN8 | ✓ | 8 | 31.6471 | **0.9788** | *29.444* | 0.9174 |

## 4. CONCLUSIONS

Remote sensing images gathered by the optical satellite are susceptible to interference by the climate that greatly reduces the usability and availability of the remotely sensed data, which is especially true when it comes to cloud covers. Hence, cloud removal is an inevitable and unescapable preprocessing procedure before image analysis and utilization. Even though the neural networks with the attention mechanism have shown their tremendous potential in substantial computer vision and natural language processing tasks, but the relevant researches and applications for remote sensing satellite cloud removal are still relatively finite. This paper is the first that introduces the attention mechanism with the generative adversarial network into the remote sensing imagery cloud removal task, as well as the first to proposed a full-attention framework (MLA-GAN). The multi-head linear attention mechanisms in each layer of the encoder enable MLA-GAN to capture and utilize the long-range dependencies and relationships between overall pixels in feature maps. Compare with traditional methods, CNN-based methods, and GAN-based methods, the experiments on RICE1 and RICE2 datasets verified the best cloud removal ability of the proposed framework.